\documentclass[runningheads]{llncs}

\usepackage[arxiv,year=2026,ID=6566]{eccv}

\usepackage{eccvabbrv}

\usepackage{graphicx}
\usepackage{booktabs}

\usepackage[accsupp]{axessibility}  
\usepackage{tikz} 
\usetikzlibrary{arrows.meta}   
\usetikzlibrary{positioning}   
\usetikzlibrary{fit}           
\usetikzlibrary{backgrounds}   

\usepackage{hyperref}

\usepackage{orcidlink}

\begin{document}

\title{High-Fidelity 4D Hand-Object Capture via Multi-View Spatiotemporal Tracking and Physics-Aware Gaussians}

\titlerunning{High-Fidelity 4D Hand-Object Capture}

\author{Bo Peng\inst{1,2}\textsuperscript{*} \and
Xu Chen\inst{1} \and
Yi Gu\inst{1,3} \and
Hidenobu Matsuki\inst{1} \and
Mingsong Dou\inst{1} \and
Jingjing Shen\inst{1} \and
Deying Kong\inst{1} \and
Juyong Zhang\inst{2} \and
Zhengyang Shen\inst{1}\textsuperscript{*}}

\authorrunning{B.~Peng et al.}

\institute{Google XR \and
University of Science and Technology of China (USTC) \and
The Hong Kong University of Science and Technology (Guangzhou)\\
\textsuperscript{*}Equal contribution}

\maketitle

\begin{abstract}
The growing demand for high-fidelity 4D hand-object interaction (HOI) data in embodied AI and spatial computing is currently bottlenecked by the reliance on pre-scanned object templates and physical markers. 
While recent methods have demonstrated promising results in reconstructing 4D hand-object interaction from videos, they are highly sensitive to initial estimates of hand and object poses. 
Yet, estimating these poses from images is challenging, in particular under severe occlusion which is inherent in hand-object interaction scenarios.
We propose a novel system for the robust and accurate reconstruction of hands and objects from synchronized and calibrated multi-view videos without requiring any templates or markers.
Our system consists of two main components with key innovations: (1) a multi-view feed-forward transformer model that aggregates cross-view geometry and temporal cues to provide a reliable, metric-consistent initialization for both poses and dense object geometry,
and (2) a hand-object physics-aware Gaussian-based optimization framework to refine the initial estimates, integrating tetrahedral constraints, collision refinement, and appearance decomposition to produce physically plausible and visually accurate reconstruction.
Validated on public benchmarks and an extensive internal dataset, our pipeline achieves highly robust, artifact-free reconstruction, providing an efficient foundation for automated 4D asset generation.
Our project page are available at \url{https://hostpg.github.io/}.

\end{abstract}

\section{Introduction}
\label{sec:intro}

The rapid evolution of embodied AI and immersive spatial computing has catalyzed a surging demand for high-fidelity 4D hand-object interaction (HOI) data~\cite{chao2021dexycb,fan2023arctic,fu2025gigahands,grauman2022ego4d,taheri2020grab,hampali2020honnotate,hampali2022keypointtransformer, yang2022oakink,damen2018scaling}.
Conventionally, capturing such data requires known object templates and physical markers~\cite{fan2023arctic} or manually annotated keypoints~\cite{chao2021dexycb} to track the object and hand poses.
This strict dependency prevents these methods from generalizing to diverse, unknown objects in the wild, severely bottlenecking the \textcolor{black}{data production throughput} of existing capture pipelines.

Several recent methods have attempted to reconstruct 4D interacting hands and objects from video alone~\cite{cong2025dytact,fu2025gigahands,pokhariya2024manus,fan2024hold}, bypassing markers and templates by optimizing 3D Gaussian or NeRF representations.
While promising, these methods are highly sensitive to initialization.
An inaccurate estimate of initial hand or object poses often leads the optimization toward poor local minima or total divergence.
Despite rapid progress~\cite{yu2025dyn,chen2025hort,wang2025unihope, sam3dteam2025sam3d3dfyimages}, estimating hand and/or object poses from images remains challenging due to inherent depth ambiguity and severe dynamic occlusions.
Consequently, existing video-based capture methods often lack robustness or must impose restrictive assumptions, such as requiring objects to remain static~\cite{pokhariya2024manus} or move at low velocities~\cite{cong2025dytact}.


To bridge this gap, we propose a novel system capable of reconstructing hands and unknown objects during natural interactions from \textcolor{black}{multi-view video captured by a calibrated, 360-degree synchronized camera rig (comprising 21 cameras at 30 fps)} with significantly improved robustness and accuracy.
Our framework adopts a two-stage design with key innovations at each phase.
We first employ a novel multi-view, spatiotemporal feed-forward model for robust initialization, followed by a physics-aware Gaussian-based optimization framework that refines geometry and appearance. 

Our first innovation addresses the initialization bottleneck via the hand object spatiotemporal transformer (HOST).
Existing hand or object pose estimators~\cite{yu2025dyn,chen2025hort,wang2025unihope, sam3dteam2025sam3d3dfyimages} are primarily designed for monocular input and rely heavily on learned priors, which \textcolor{black}{often fail to resolve the severe depth ambiguities and occlusions inherent in complex interactions}.
While recent foundational 3D models (e.g., VGGT~\cite{wang2025vggt}) effectively leverage multi-view cues, they are generally tailored for reconstruction of static scenes.
HOST extends these foundational models to the task of multi-view tracking of dynamic interacting hands and objects.
By introducing a factorized spatiotemporal attention mechanism, HOST efficiently aggregates information across viewpoints and frames, producing accurate and temporally coherent object and hand pose estimates. 
In addition, even if the object is partially occluded in some frames, HOST is able to reconstruct a complete and dense point cloud by using the geometric cues in all frames and viewpoints.
HOST effectively resolves depth ambiguities where monocular trackers fail, ensuring metric 3D consistency and providing a reliable geometric foundation for subsequent refinement.

Our second innovation is a hand-object physics-aware Gaussian (HOPG) optimization framework, which elevates this initialization into high-fidelity 4D assets.
We model the two hands and the unknown object with independent 2D Gaussian~\cite{huang20242d} clouds, which are initialized with the tracked pose parameters as well as the reconstructed dense point cloud from HOST.
We optimize the 2D Gaussians along with the object and hand poses by minimizing the photometric loss.
To prevent implausible distortions or volume collapse of the hand during close interactions~\cite{cong2025dytact, peng2025pica}, we anchor the hand Gaussian on a tetrahedral mesh proxy and regularize the optimization with a volumetric As-Rigid-As-Possible (ARAP) energy~\cite{sorkine2007rigid}. 
Furthermore, we incorporate a dedicated physics-aware collision resolution step to ensure no penetration.
To reconstruct high-fidelity appearance of hand and object, we further introduce a temporal batching strategy and a decoupled model~\cite{saito2024relightable} that disentangles view-independent albedo and view-dependent specular.
This decomposition prevents environmental lighting from being ``baked'' into the geometry, recovering the intrinsic surface colors.

Leveraging our framework, we processed a large-scale dataset of 77 complex sequences involving 32 unique objects with complex geometries.
Achieving an automated success rate exceeding 93\%, we demonstrate our system's exceptional robustness for \textcolor{black}{automated 4D asset generation}.
We extensively evaluate both our isolated modules (HOST and HOPG) and the full pipeline across a public benchmark~\cite{moon2020interhand2}, the VEPHand dataset~\cite{anonymous_handbooth}, and our custom dataset. 
The results demonstrate the effectiveness of our framework across three core tasks: it establishes superior robust pose tracking, excels in dynamic novel view synthesis, and recovers highly accurate dense geometry.

In summary, our key contributions are:
\begin{itemize}
\item A system that robustly and accurately reconstructs hand object interaction from \textcolor{black}{synchronized multi-view videos (e.g., using a 21-camera 360-degree rig)} without requiring pre-scanned templates or markers, \textcolor{black}{automating the pipeline for efficient data acquisition}.
\item A spatiotemporal transformer model that \textcolor{black}{explicitly leverages cross-view and temporal cues}, providing a reliable geometric initialization even under severe occlusion.
\item A physics-aware Gaussian optimization framework integrating tetrahedral constraints, collision refinement, and appearance decomposition to produce physically plausible and visually accurate 4D reconstructions.
\end{itemize}

\section{Related Works}
\label{sec:related_work}
\subsubsection{Neural Hand and HOI Rendering.}
Reconstructing high-fidelity dynamic hand-object interactions (HOI) is a long-standing challenge. 
To address this, foundational work first focused on bare-hand neural rendering. 
Early neural implicit methods, such as HandAvatar~\cite{chen2023handavatar}, achieved high fidelity by disentangling hand geometry and texture but suffered from slow rendering speeds. 
While subsequent works like LiveHand~\cite{mundra2023livehand} overcame this to achieve real-time performance via mesh-guided sampling, 3DGS has recently emerged as a fundamentally more efficient alternative.
GauHuman~\cite{hu2024gauhuman} models articulated humans by warping 3D Gaussians via Linear Blend Skinning, and HandSCS~\cite{dong2025handscs} explicitly handles severe hand articulations by introducing a Structural Coordinate Space to bridge sparse skeletal joints and dense Gaussians.

Building upon these bare-hand reconstruction foundations~\cite{guo2025handnerf++, ivashechkin2025handocc, dong2025handscs}, a substantial body of work~\cite{ye2022s, xie2025cari4d} tackles HOI through per-scene optimization to recover precise surfaces and contact mechanics. 
Recent methods achieve this using articulated 3D Gaussians (MANUS~\cite{pokhariya2024manus}, DyTact~\cite{cong2025dytact}), compositional implicit fields (HOLD~\cite{fan2024hold}), or generative diffusion priors for unobserved regions~\cite{zhenyuan2025bigs, liugeneralizable, wang2025magichoi}. 
However, these optimization frameworks suffer from two critical limitations: they inevitably collapse when their required tracking initializations drift under severe occlusions, and they frequently ``bake'' environmental illumination into the geometry. \textcolor{black}{To bypass the fragility of traditional tracking bottlenecks, recent advances like EasyHOI~\cite{liu2025easyhoi} leverage large vision-language models to directly estimate hand-object geometry from monocular in-the-wild inputs before executing optimization.}

\subsubsection{Hand-Object Pose Tracking.}
To provide the tracking priors required by the aforementioned optimization pipelines, an emerging line of inference-based methods trains feed-forward models to directly regress hand and object states from images. 
For instance, HORT~\cite{chen2025hort} employs a coarse-to-fine transformer architecture to efficiently reconstruct dense 3D point clouds of hand-held objects and their hand-relative poses. 
For complex hand motions, Dyn-HaMR~\cite{yu2025dyn} recovers global 4D trajectories from monocular videos by combining hierarchical hand tracking with generative motion priors and SLAM. \textcolor{black}{To capture hands across alternative viewpoints, multi-view frameworks have also been explored; notably, POEM~\cite{yang2025multi} introduces a point-embedded transformer for single-frame multi-view hand reconstruction.} Additionally, foundation models like SAM 3D~\cite{chen2025sam} have generalized single-image 3D perception, demonstrating remarkable capabilities in predicting object geometry, texture, and scene layout directly from in-the-wild 2D images. 

However, \textcolor{black}{monocular methods remain inherently susceptible to severe depth ambiguities, while single-frame multi-view fusion methods often fail to maintain temporal consistency.} 
In contrast, our HOST module \textcolor{black}{explicitly} aggregates \textcolor{black}{continuous} cross-view and temporal cues to resolve these ambiguities, establishing a metrically reliable \textcolor{black}{and temporally coherent} initialization for the downstream optimization.

\subsubsection{Feed-Forward Geometric Estimation.}
The success of large-scale pre-training has shifted 3D geometric estimation toward feed-forward regression~\cite{hong2023lrm,wang2025pi}. 
Pioneered by DUSt3R~\cite{wang2024dust3r, leroy2024grounding, yang2025fast3r}, foundational works demonstrate that dense 3D point maps can be directly regressed from unposed images. 
Building upon this, VGGT~\cite{wang2025vggt} and its successors~\cite{shen2025fastvggt, wu20254dlangvggt} introduced large networks~\cite{dosovitskiy2020image} for joint camera and scene geometry prediction. 
To adapt these spatial priors for dynamic 4D environments without incurring prohibitive computational overhead, recent works like 4D-VGGT~\cite{wang20254d} and MonST3R~\cite{zhang2024monst3r} introduce spatiotemporal decoupling to model structural consistency and motion continuity separately. 
\textcolor{black}{Concurrently, drawing inspiration from these visual-geometry grounded architectures, HGGT~\cite{liu2026hggt} jointly infers 3D hand meshes and camera poses from uncalibrated arbitrary views.}
Unlike these foundational \textcolor{black}{or single-entity} models, which primarily output generic scene point clouds \textcolor{black}{or focus exclusively on bare hands}, HOST is explicitly designed to parse complex \textcolor{black}{hand-object} interactions. By introducing a spatiotemporal attention mechanism, we efficiently aggregate multi-view and temporal cues to jointly regress hand and object states and entity-specific dense geometry. 
This effectively translates powerful generic 3D priors into a specialized articulated tracker, providing a reliable 4D initialization for our downstream HOPG pipeline.

\section{Method}
\label{sec:method}
\begin{figure}[t]
    \centering
    \includegraphics[width=\textwidth]{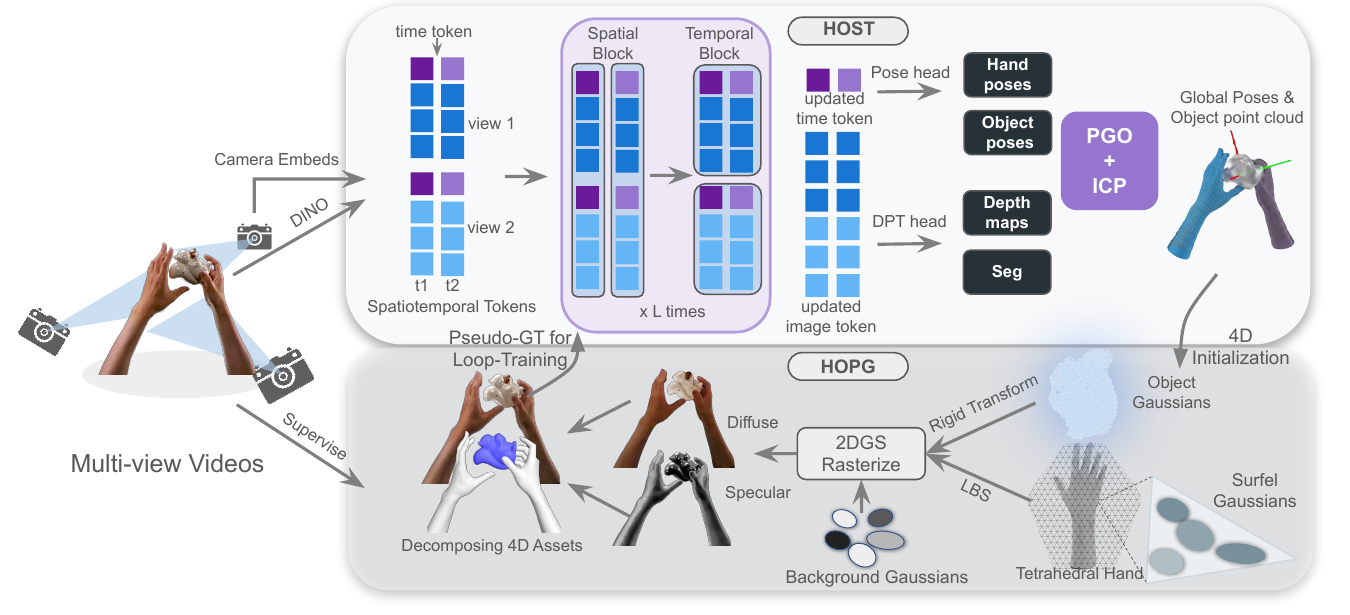}
    \vspace{-7mm}
    \caption{\textbf{Overview of our framework.} Our pipeline operates in two stages. First, the Hand Object Spatiotemporal Transformer (HOST) processes multi-view videos to robustly regress parametric hand/object poses, dense object point clouds, and segmentation masks, yielding a metric 3D initialization. 
    Second, the Hand Object Physics-aware Gaussian (HOPG) module leverages this initialization to optimize a hybrid 2D Gaussian representation. 
    By enforcing structural constraints and explicitly decoupling diffuse and specular appearance, HOPG effectively prevents illumination bake-in, reconstructing high-fidelity 4D assets.}
    \label{fig:pipeline}
\end{figure}

Given synchronized multi-view RGB videos and calibrated camera parameters, our framework first employs the Hand Object Spatiotemporal Transformer  (Sec.~\ref{sec:hot}) to establish a robust multi-view spatiotemporal prior, followed by a Hand Object Physics-aware Gaussian (HOPG) refiner (Sec.~\ref{sec:HOPG}). The full pipeline is illustrated in Fig.~\ref{fig:pipeline}.




\subsection{HOST: Hand-Object Spatiotemporal Transformer}
\label{sec:hot}

The first stage of our framework establishes a robust geometric prior to guide the downstream volumetric refinement. 
Given a sequence of $F$ frames across $V$ synchronized static views, HOST regresses temporally coherent 4D attributes and dense geometric features.  
%
Formally, HOST defines a mapping $\mathcal{F}$ as follows:
\begin{equation}
\left( \mathbf{D}_{1:F, 1:V}, \mathbf{\Theta}^h_{1:F}, \mathbf{\Theta}^o_{1:F} \right) = \mathcal{F}\left( \mathbf{I}_{1:F, 1:V}, t_{1:F} , \mathbf{E}_{1:V}, \mathbf{K}_{1:V} \right),
\end{equation}
where $\mathbf{D}_{1:F, 1:V}$ denote the dense predictions (i.e., semantic segmentation masks, depth maps, and point maps) across all $F$ frames and $V$ viewpoints; $\mathbf{\Theta}^h_{1:F}$ and $\mathbf{\Theta}^o_{1:F}$ refer to the full sequences of hand and object parameters, respectively; $\mathbf{I}_{1:F, 1:V}$ denotes the input multi-view video sequence; $\mathbf{E}_{1:V}$ and $\mathbf{K}_{1:V}$ denote the camera extrinsic and intrinsic parameters for all views, and $t_{1:F}$ represent the relative frame identifiers.

\noindent\textbf{Spatiotemporal Feature Backbone.}
HOST builds upon the VGGT~\cite{wang2025vggt} transformer backbone but extends it to a spatial and temporal structure. 
%
To encode spatial and temporal information, each image $\mathbf{I}_{f,v}$ is patchified into tokens $\mathbf{T}_{\text{image}}$ using a frozen DINOv2~\cite{oquab2023dinov2} backbone, and augmented with view-consistent, ray-based camera embeddings~\cite{keetha2025mapanything}. Simultaneously, relative frame identifiers $t_{f}$ are mapped to view-shared time tokens $\mathbf{T}_{\text{time}}$, which are then concatenated with $\mathbf{T}_{\text{image}}$ to form the final input sequence $(\mathbf{T}_{\text{time}}, \mathbf{T}_{\text{image}})$ for the factorized attention layers.

Processing massive 4D sequences ($V$ views, $F$ frames, $N$ patches) via monolithic global attention, such as the frame-global loop in VGGT, incurs a prohibitive computational complexity of $\mathcal{O}((V F N)^2)$. 
To circumvent this bottleneck while preserving rich 4D context, we propose the spatiotemporal mechanism. 
By retaining foundational intra-frame attention but explicitly factorizing the intractable global attention into alternating Spatial and Temporal Blocks (Fig.~\ref{fig:pipeline}), the feature update cycle for the $k$-th group is formulated as:
\begin{align}
\mathcal{B}_{\text{spatial}} &: \mathbf{T'}_{k} = \mathcal{A}_{\text{view}}\left(\mathcal{A}_{\text{frame}}(\mathbf{T}_{k})\right), \\
 \mathcal{B}_{\text{temporal}} &: \mathbf{T}_{k+1}=\mathcal{A}_{\text{temporal}}\left(\mathcal{A}_{\text{frame}}(\mathbf{T'}_{k})\right),
\end{align}
Here, $\mathcal{A}_{\text{frame}}$ performs local intra-frame self-attention (costing $\mathcal{O}(V F N^2)$). The Spatial Block ($\mathcal{B}_{\text{spatial}}$) aggregates stereo cues via cross-view attention $\mathcal{A}_{\text{view}}$ for each frame independently, bounding its cost to $\mathcal{O}(F(V N)^2)$. Conversely, the Temporal Block ($\mathcal{B}_{\text{temporal}}$) aggregates motion cues via temporal attention $\mathcal{A}_{\text{temp}}$ for each view independently, costing $\mathcal{O}(V(F N)^2)$. This factorized design successfully aggregates global 4D context while drastically reducing the quadratic computational overhead. In practice, we train the framework using a chunking strategy, e.g.,  a typical chunk we use in training is 5 frames with 6 views.

\noindent\textbf{Dense Geometry Heads.} 
We employ a Dense Prediction Transformer (DPT) architecture~\cite{ranftl2021vision} to decode the features into a set of per frame and per view dense maps $\mathbf{D}$:
multi-class probability map for the left hand, right hand, and the interacting object; depth map and point map.
This joint supervision encourages the network to disentangle the tightly interacting entities implicitly, providing a clean, entity-specific geometric foundation.

\noindent\textbf{Global Kinematics Heads.} 
To estimate the global 4D kinematic state, we pass the final stage time tokens $\mathbf{\hat{T}}_{\text{time}}$, which encode the overall spatiotemporal dynamics, to dedicated MLP heads that regress the per-frame kinematic states for hand and object parameters. The hand parameters $\mathbf{\Theta}^h=\{ \alpha, \beta \}$ here refer to the identity $\alpha$  and poses $\beta$ w.r.t. a given parametric hand model~\cite{MANO:SIGGRAPHASIA:2017}. 

For the rigid object transform, we adopt a hybrid state representation comprising the absolute centroid translation $\mathbf{t}^{\text{abs}}_f$ in world coordinates and the relative rotation $\Delta \mathbf{R}_{1 \to f}$ (with respect to the first frame of the input chunk).
%
This asymmetric design explicitly addresses the distinct observability of translation and rotation. Because the canonical orientation of an unseen object is inherently ambiguous without a predefined template, predicting relative rotation is strictly necessary. While it may seem intuitive to also predict relative translations, integrating frame-to-frame offsets inevitably accumulates trajectory drift over time. 
Instead, since the object's absolute 3D geometric center remains continuously observable from spatiotemporal cues, we directly regress its absolute translation. 
This firmly anchors the spatial trajectory, eliminating long-term positional drift and confining any unavoidable error accumulation strictly to the rotational domain.

\noindent\textbf{Training Losses.}
\label{sec:sthot_loss}
We follow the VGGT ~\cite{wang2025vggt} for dense prediction (segmentation, depth and point map) training and apply geodesic distance~\cite{hartley2013rotation, huynh2009metrics} on rotation and L2 losses on translation and registered meshes from forward kinematics.

\subsubsection{Inference.}
\label{sec:hot_inference}
To process arbitrarily long videos under memory constraints, we split sequences into overlapping chunks using a sliding window with a stride of one. For hands, simply averaging the overlapping predictions provides sufficient temporal smoothing. For rigid objects, however, we prevent long-term trajectory drift by aggregating chunk-wise predictions in two steps:


We first employ \textit{Pose Graph Optimization (PGO)}~\cite{kummerle2011g,dellaert2012factor} to combine per-window pose estimates into a continuous trajectory. 
Specifically, we optimize absolute object pose estimate $\mathbf{\Theta}^o = (\mathbf{R}, \mathbf{t})$ with the predicted relative rotation edges $\mathcal{E}$ and absolute translations:
\begin{equation}
    \mathop{\arg\min}_{\{\mathbf{R}_f, \mathbf{t}_f\}} \sum_{(i,j) \in \mathcal{E}} \left\| \mathbf{R}_i^T \mathbf{R}_j - \Delta \mathbf{R}_{i \to j} \right\|_F^2 + \lambda_{\text{abs}} \sum_{f} \left\| \mathbf{t}_f - \mathbf{t}^{\text{abs}}_f \right\|^2. 
\end{equation}

To eliminate residual drift, we then unproject the masked depth maps into 3D point clouds and align them via frame-to-model ICP~\cite{besl1992method,zhou2016fast} with the PGO trajectory as a robust initialization. 
In this way, we are able to infer accurate object pose trajectories even in long videos. Fusing these aligned observations yields the canonical object point cloud $\mathbf{V}_{\text{obj}}^{\text{cano}}$.

In summary, this inference strategy aggregates per-window predictions into globally consistent hand parameters $\mathbf{\Theta}^h$, absolute object poses $\mathbf{\Theta}^o$, and an optional canonical point cloud of the object $\mathbf{V}_{\text{obj}}^{\text{cano}}$. These outputs are then used to initialize the subsequent HOPG refinement.

\subsection{HOPG: Hand-Object Physics-aware Gaussian Optimization }
\label{sec:HOPG}

In this stage, we jointly optimize three distinct 2D Gaussian fields, hand, object, and background, leveraging the pose and shape estimates from HOST as an initialization.



\noindent\textbf{Volumetric Hand.} 
We utilize a volumetric deformation model inspired by VEPHand~\cite{anonymous_handbooth} as our geometric proxy.
To faithfully model \textcolor{black}{anatomical thickness} and to resolve collision during close interactions, we embed the hand surface within a canonical tetrahedral mesh.
However, rather than optimizing per-frame volumetric variables, we predict these intrinsic tetrahedral deformations dynamically via a small MLP network. 
 
Specifically, let $\mathbf{v}_c$ denote the canonical surface vertices and $\boldsymbol{\rho}_c$ represent the vertices of the underlying canonical tetrahedral mesh. 
Unlike standard parametric models that apply Linear Blend Skinning (LBS) directly to the base surface, our volumetric model first applies non-rigid, pose-dependent deformations in the tetrahedral domain:
\begin{align}
\text{(Standard Surface)} \quad & \mathbf{v}_p = \text{LBS}(\mathbf{v}_c, \alpha, \beta), \\
\text{(Our Volumetric)} 
\quad & \boldsymbol{\rho}'_c = \boldsymbol{\rho}_c + \text{MLP}(\boldsymbol{\rho}_c, \beta), \\
& \mathbf{v}'_c = \mathcal{M}_{\text{tet} \to \text{surf}}(\boldsymbol{\rho}'_c), \\
& \mathbf{v}_p = \text{LBS}(\mathbf{v}'_c, \alpha, \beta),
\end{align}
where $\alpha$ and $\beta$ represent the identity shape and kinematic pose parameters, respectively. In our formulation, a lightweight MLP predicts pose-dependent offsets for the tetrahedral control points to yield the deformed volume $\boldsymbol{\rho}'_c$. We then compute the updated canonical surface vertices $\mathbf{v}'_c$ via a predefined barycentric mapping operator $\mathcal{M}_{\text{tet} \to \text{surf}}$. Geometrically, each surface vertex is initialized inside a specific tetrahedron. This operator calculates the deformed surface position simply by interpolating the displaced tetrahedral vertices $\boldsymbol{\rho}'_c$ using fixed barycentric weights. Consequently, the surface acts as a continuous skin strictly bound to the underlying volumetric flesh. Finally, standard Linear Blend Skinning (LBS) articulates these volume-aware canonical surface vertices into the final posed space $\mathbf{v}_p$.

 
Following PICA~\cite{peng2025pica}, we anchor 2D Gaussians to this deformed mesh using barycentric interpolation, ensuring the radiance field strictly inherits the kinematic consistency of the hand model.


\noindent\textbf{Rigid Object.} 
The object is represented by a set of canonical 2D Gaussians, seamlessly initialized from the fused canonical point cloud $\mathbf{V}_{\text{obj}}^{\text{cano}}$ provided by HOST. During refinement, we jointly optimize these canonical Gaussian attributes and the per-frame rigid transformations. 

\noindent\textbf{Background.} 
We model the static environment using a separate set of 2D Gaussians initialized on a large enclosing sphere. 


\subsubsection{Appearance Model.}
\label{sec:appearance}
Standard Gaussian Splatting~\cite{kerbl20233d} bakes view-dependent lighting into the spherical harmonics (SH) of each primitive. In highly dynamic hand-object interactions, this entanglement between extreme rotations and illumination causes severe flickering and texture baking. To reconstruct physically and temporally consistent assets, we explicitly factorize intrinsic albedo, diffuse shading, and specularity. Following Saito et al.~\cite{saito2024relightable}, the dynamic diffuse color $\mathbf{C}_{\text{diff}}$ of a Gaussian is formulated by convolving environmental light SH coefficients $\mathbf{L}$ with dynamically rotated canonical transfer coefficients $\mathbf{d}(\mathbf{R})$, modulated by a learnable view-independent albedo $\mathbf{a}$:
\begin{equation}
\mathbf{C}_{\text{diff}} = \mathbf{a} \odot \sum_{i=1}^{(n+1)^2} \mathbf{L}_i \mathbf{d}_i(\mathbf{R}),
\end{equation}
where $n$ is the SH degree, and $\mathbf{R}$ defines the Gaussian's rotation from canonical to world space. Explicitly applying $\mathbf{R}$ ensures the transfer coefficients remain aligned with the global environment light, effectively decoupling illumination from the primitive's motion. Furthermore, to prevent view-dependent highlights from polluting this intrinsic albedo, we model specularity using a dedicated lightweight MLP~\cite{verbin2024ref}. 
The network processes the Integrated Directional Encoding (IDE) of the reflection vector $\mathbf{r}$ and the Gaussian's learnable surface roughness $\rho$ to output a specular residual. 
The final composited color is then evaluated as $\mathbf{C}_{\text{total}} = \mathbf{C}_{\text{diff}} + s \cdot \text{MLP}_{\text{spec}}(\text{IDE}(\mathbf{r}, \rho))$, where $s$ is a learnable, primitive-specific specular intensity.

\noindent\textbf{Losses.} 
The data terms are driven by standard photometric losses (L1, D-SSIM, LPIPS). 
For physical fidelity, we impose an As-Rigid-As-Possible (ARAP) energy~\cite{sorkine2007rigid} on the underlying tetrahedral mesh for volume preservation on deformed canonical tetrahedra $\mathbf{\rho'_c}$ and a ray-tracing based collision loss~\cite{anonymous_handbooth} on posed surface mesh $\mathbf{v}_p$ to prevent both inter and self penetration.

\begin{figure}[t]
    \centering
    \includegraphics[width=\textwidth]{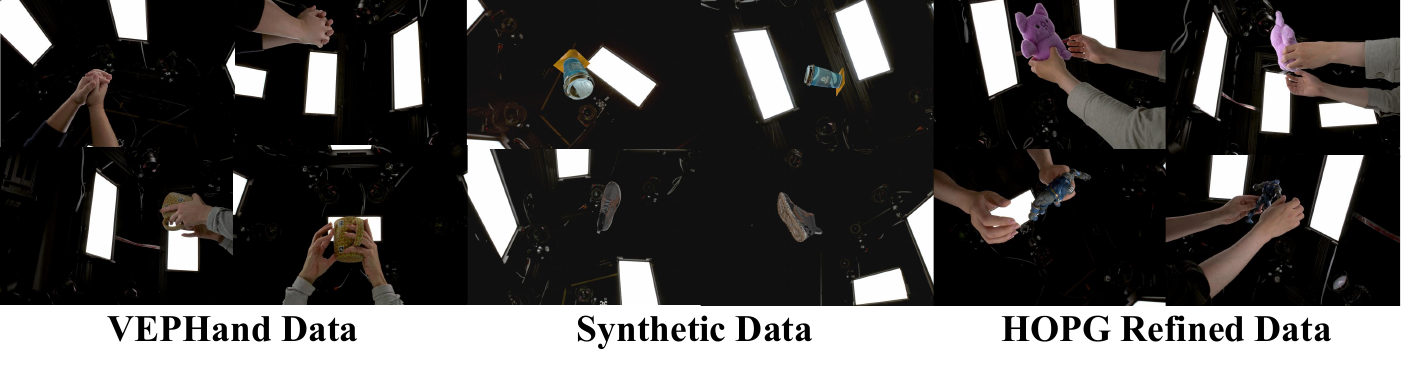}
    \vspace{-8mm}
    \caption{\textbf{Overview of our composite training dataset.} \textbf{Left:} Multi-view captures from the VEPHand~\cite{anonymous_handbooth} dataset. \textbf{Middle:} Synthetic data rendered in Blender using diverse textured objects from Texverse~\cite{zhang2025texverse} Dataset. \textbf{Right:} Custom captures refined via our HOPG module.}
    \label{fig:dataset}
    \vspace{-7mm}
\end{figure}

\subsection{Dataset Roadmap}
\label{sec:dataset}
To ensure kinematic diversity and geometric generalizability, HOST is trained on a comprehensive composite dataset.
The foundation comprises 1,600 multi-view sequences from VEPHand Dataset~\cite{anonymous_handbooth}, featuring high-quality scans and hand-object registrations across 17 3D-printed CAD models. 
To overcome the limited object diversity of this set, we synthesize a supplementary dataset in Blender, rendering 500 high-resolution textured objects~\cite{zhang2025texverse} simulated with randomized 6D trajectories. 
Finally, to capture real-world complexity, we collect a custom dataset of 47 everyday objects augmented with miniature ArUco markers to reliably acquire initial relative poses. 
These captures are then processed by our HOPG refiner. 
Once converged, HOPG acts as a robust pseudo-label generator, yielding refined ground-truth equivalents (cleaner masks, accurate depth, and collision-free poses). These high-fidelity annotations are then fed back to fine-tune HOST, progressively closing the domain gap without manual intervention.

\section{Experiment}
\label{sec:exp}
To comprehensively evaluate our framework's capabilities in photorealistic novel view synthesis (NVS), physically plausible geometry recovery, and industrial-level scalability, we conduct extensive experiments on a composite evaluation suite comprising both public benchmarks and diverse in-house captures.
\subsection{Experimental Setup}
\begin{figure*}[ht]
\vspace{-9mm}
    \centering
    \includegraphics[width=\linewidth]{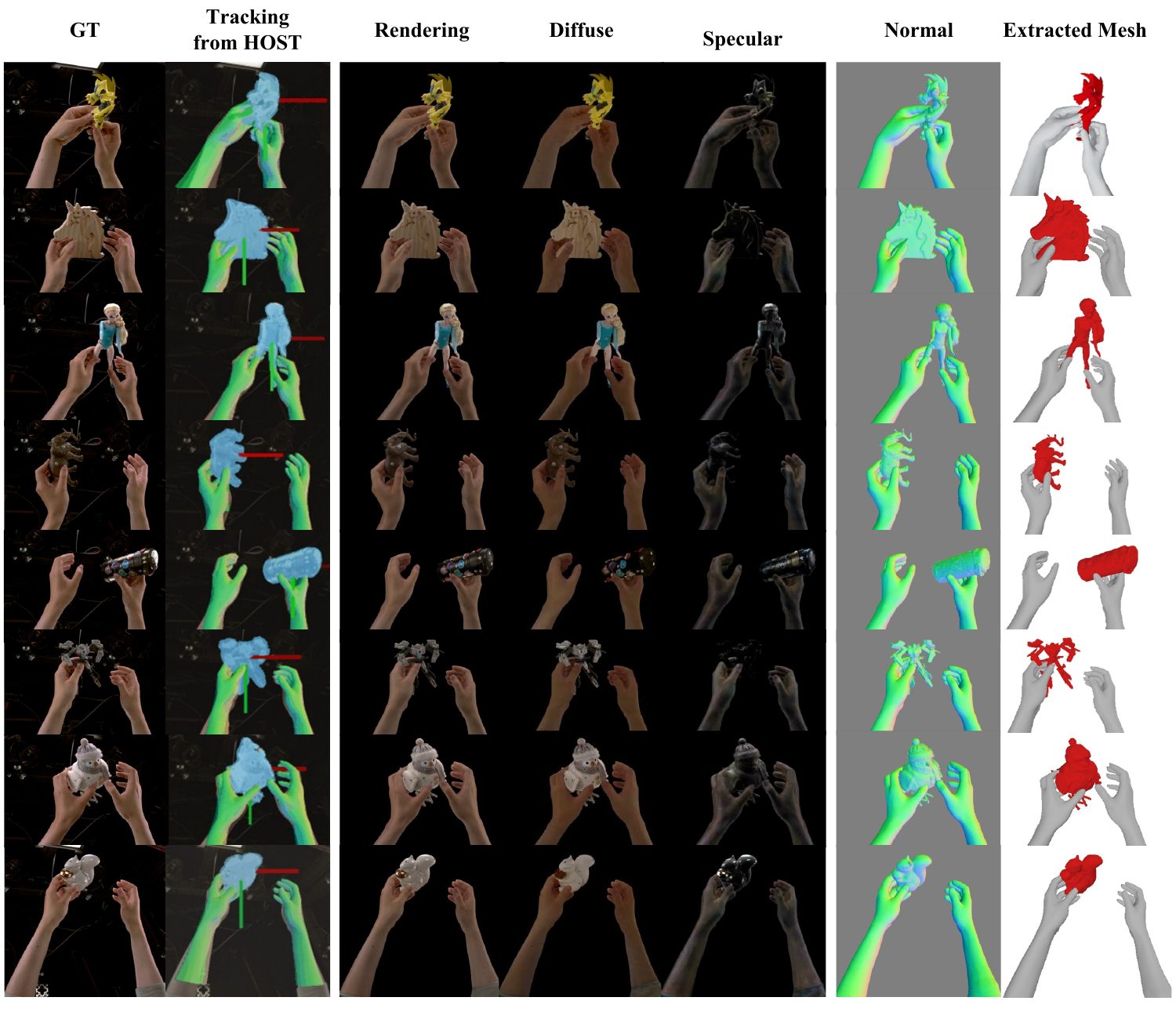}
    \vspace{-7mm}
    \caption{\textbf{Automated, Scalable High-Fidelity Capture across Diverse Objects.} Given calibrated multi-view images (GT), HOST provides robust tracking initializations. HOPG then achieves photorealistic rendering by decomposing appearance into diffuse and specular components. This explicit lighting decoupling prevents artifacts from baking into the geometry, yielding clean normal maps and high-fidelity 3D meshes.}
    \label{fig:main_gallery}
    
\end{figure*}

\noindent\textbf{Implementation Details}
HOST is built on a ViT-Large backbone with frozen DINOv2~\cite{oquab2023dinov2} patch embeddings and initialized with pre-trained VGGT~\cite{wang2025vggt} weights. 
We optimize the network using AdamW in bfloat16 precision. 
To stabilize early training, we introduce a 2,000-step geometry warmup where we only optimize the backbone and dense geometry heads. 
The model is first pre-trained for 200k steps on 64 TPU v5 accelerators ($lr=8.0\times 10^{-5}$) on VEPHand data, and subsequently fine-tuned for 20k steps on 8 NVIDIA H100 GPUs ($lr=1.0\times 10^{-5}$) on synthetic and our HOPG refined data. 
For the HOPG refinement, we use a batch size of 5 frames to explicitly enforce the decoupled appearance model. 
A typical sequence of 21 views and 600 frames takes about 30k iterations (roughly 5 hours) to converge on a single NVIDIA A100 GPU.

\begin{figure}[t]
    \centering
    \includegraphics[width=\textwidth]{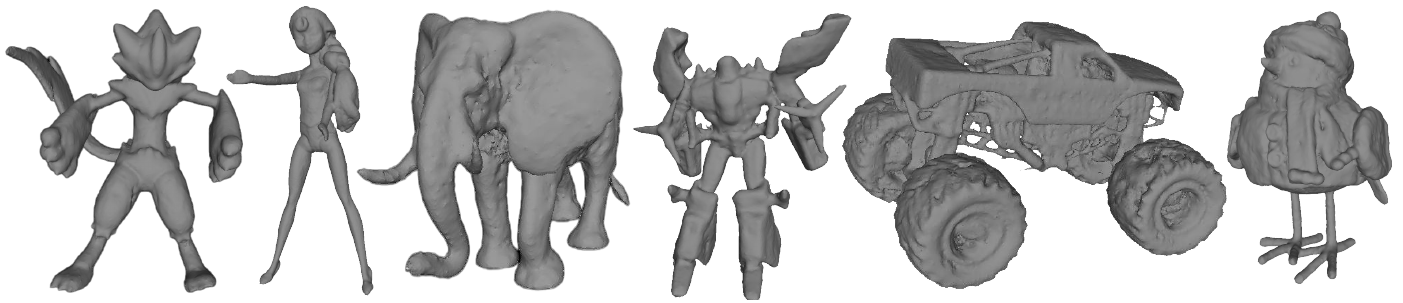}
    \vspace{-8.5mm}
    \caption{\textbf{High-Fidelity Object Geometry.} Detailed 3D meshes of complex, real-world objects extracted via TSDF fusion.}
    \label{fig:mesh}
    \vspace{-1mm}
\end{figure}

\noindent\textbf{Scalability Validation.} 
To validate scalability, we deploy our pipeline on a newly captured multi-view dataset of diverse objects. 
As qualitatively demonstrated in Fig.~\ref{fig:main_gallery}, our framework operates entirely automatically, successfully reconstructing 72 out of 77 interaction sequences across 32 unique objects, thereby achieving an exceptional $\sim$93.5\% success rate. 
Furthermore, we also showcase the high-resolution 3D object meshes extracted via TSDF fusion~\cite{curless1996volumetric} in Fig.~\ref{fig:mesh}. 
The few failures stem from challenging object properties, such as geometrically thin structures (pens) and highly symmetric, textureless surfaces (tapes), which induce severe pose ambiguities during HOST initialization.

\noindent\textbf{Evaluation Datasets.} 
We comprehensively evaluate our framework across three distinct domains. First, a representative 10-sequence subset from our custom captures serves as the primary testbed for overall pipeline comparisons. Second, we benchmark our isolated HOPG module on InterHand2.6M~\cite{moon2020interhand2}. 
Finally, to rigorously evaluate both our feed-forward initialization and the final dense geometry reconstruction, we utilize the VEPHand Dataset~\cite{anonymous_handbooth}, which provides high-quality per-frame volumetric pseudo-ground truth across 7 complex sequences.

\subsection{Novel View Synthesis}
In this section, we first evaluate novel view synthesis capability of HOPG on InterHand2.6M and then assess the full pipeline on our Custom Dataset.

\begin{figure}[t]
    \centering
    \includegraphics[width=\textwidth]{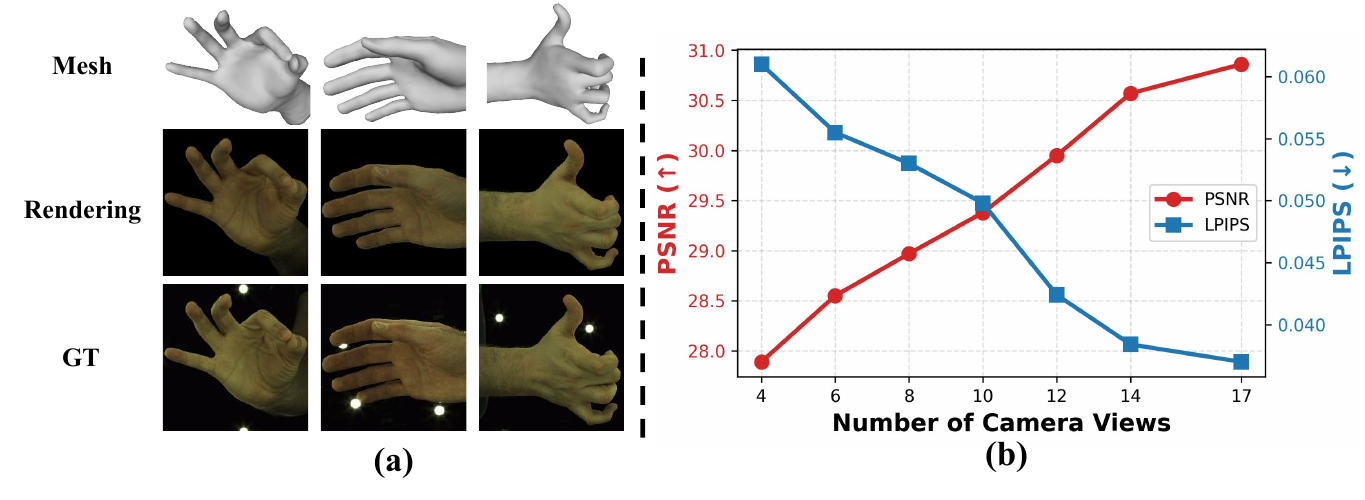}
    \vspace{-9mm}
    \caption{\textbf{InterHand2.6M Validation and View Robustness.} \textbf{(a)} Our HOPG module achieves high-fidelity rendering and accurate geometry on bare-hand benchmarks. \textbf{(b)} End-to-end reconstruction remains highly stable across varying numbers of input cameras, maintaining strong accuracy even under extreme sparsity (4 views).}
    \label{fig:hand26m_camera}
    \vspace{-4mm}
\end{figure}

\noindent\textbf{HOPG on InterHand2.6M.} 
We follow the HandSCS~\cite{dong2025handscs} protocol (10 training, 15 held-out views) across their exact 9 selected sequences. 
As shown in Table~\ref{tab:nvs_main} (left), our method is highly competitive. 
Standard NeRf~\cite{mundra2023livehand,chen2023hand} or 3DGS baselines~\cite{dong2025handscs} artificially inflate photometric scores by relying on semi-transparent volume accumulation and baked-in illumination. 
In contrast, our formulation trades unconstrained photometric memorization for physical accuracy by enforcing strict, opaque 2D surfaces and explicitly disentangling albedo from specular components. 
We also provide qualitative rendering results in Fig.~\ref{fig:hand26m_camera}a.

\begin{figure}[h]
    \centering
    \includegraphics[width=\textwidth]{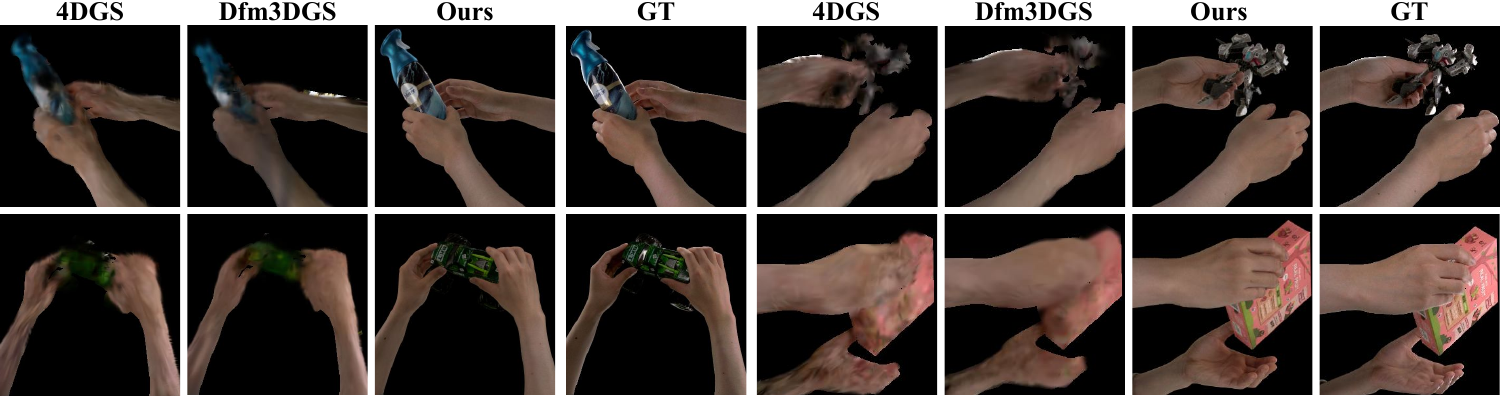}
    \vspace{-7mm}
    \caption{\textbf{Qualitative comparison on our Custom-Captured Dataset.} Our method successfully tracks difficult motions and recovers complex object geometries, whereas prior-free baselines suffer from severe motion blur and geometric collapse.}
    \label{fig:nvs_qualitative}
    \vspace{-3mm}
\end{figure}

\noindent\textbf{Full pipeline on Custom Data.}
To evaluate the complete framework, taking only raw multi-view videos and camera parameters as input, we compare against representative dynamic scene methods: 4DGS~\cite{wu20244d} and Deformable 3DGS~\cite{yang2024deformable}. 
To focus strictly on the interaction, we utilize SAM2~\cite{kirillov2023segment,ravi2024sam} to isolate the foreground hand and object. As reported in Table~\ref{tab:nvs_main} (right), our explicitly constrained representation outperforms these baselines by a massive margin ($>$\,4.5 dB in PSNR). Qualitatively (Fig.~\ref{fig:nvs_qualitative}), general dynamic methods lack structural priors, causing their optimization to degrade into severe motion blur and catastrophic geometric collapse under highly non-linear kinematics. Conversely, guided by the accurate metric priors from HOST, our framework seamlessly tracks these motions while maintaining sharp visual boundaries.

\begin{table}[t]
\centering
\caption{\textbf{Quantitative NVS results.} Comparison against specialized hand avatars on InterHand2.6M (left, 10-view train / 15-view test) and general dynamic baselines on our Custom-Captured Dataset (right, 17-view train / 4-view test).}
\label{tab:nvs_main}
\vspace{-2mm}

\begin{subtable}[t]{0.49\linewidth}
\centering
\resizebox{\linewidth}{!}{
\begin{tabular}{l c c c}
\toprule
\multicolumn{4}{c}{\textbf{InterHand2.6M}} \\
\midrule
Method & PSNR $\uparrow$ & SSIM $\uparrow$ & LPIPS $\downarrow$ \\
\midrule
LiveHand~\cite{mundra2023livehand} & 31.06 & \textbf{0.977} & 0.0384 \\
HandAvatar~\cite{chen2023hand} & 32.14 & 0.962 & 0.0437 \\
GauHuman~\cite{hu2024gauhuman} & 31.73 & 0.957 & 0.0462 \\
HandSCS~\cite{dong2025handscs} & \textbf{35.51} & 0.974 & \textbf{0.0292} \\
\midrule
\textbf{Ours} (HOPG) & 35.07 & 0.970 & 0.0348 \\
\bottomrule
\end{tabular}
}
\end{subtable}\hfill
\begin{subtable}[t]{0.49\linewidth}
\centering
\resizebox{\linewidth}{!}{
\begin{tabular}{l c c c}
\toprule
\multicolumn{4}{c}{\textbf{Custom Dataset}} \\
\midrule
Method & PSNR $\uparrow$ & SSIM $\uparrow$ & LPIPS $\downarrow$ \\
\midrule
4DGS~\cite{wu20244d} & 26.34 & 0.943 & 0.0814 \\
Dfm3DGS~\cite{yang2024deformable} & 26.32 & 0.947 & 0.0809 \\
\midrule
\textbf{Ours} (Full) & \textbf{30.86} & \textbf{0.968} & \textbf{0.0370} \\
\bottomrule
\end{tabular}
}
\end{subtable}

\vspace{-4mm}
\end{table}

\subsection{Geometry Evaluation}

To verify that our framework recovers 3D geometry rather than merely overfitting photometric projections, we evaluate it on the VEPHand Dataset with per-frame GT scan. 
We report the one-way Surface Accuracy (point-to-surface distance from the predicted vertices to the ground-truth scans, which include the forearm). 
\textcolor{black}{Importantly, as our goal is to capture the complete interaction, all reported metrics in Table~\ref{tab:geometry} are computed on the combined hand-object meshes.} 
Specifically, we compare our HOST initialization against state-of-the-art feed-forward trackers, and assess the final dense geometry refined by our full pipeline (Table~\ref{tab:geometry}).

\begin{table}[t]
    \centering
    \caption{
        \textbf{Geometry Evaluation.} One-way Surface Accuracy (mm, $\downarrow$) evaluated on \textcolor{black}{combined hand-object meshes} of the VEPHand Dataset. Our HOST module provides a superior metric initialization compared to feed-forward baselines, \textcolor{black}{including multi-view alternatives,} while our full pipeline further refines the geometry to achieve highly competitive dense accuracy. Processing time denotes the average time per frame measured on a single NVIDIA A100 GPU.
    }
    \vspace{-2mm}
    \label{tab:geometry}
    \begin{tabular}{l c c}
        \toprule
        Method & Accuracy (mm) $\downarrow$ & Processing Time $\downarrow$ \\
        \midrule
        HORT~\cite{chen2025hort} & 7.22 & 2s \\
        DynHaMR~\cite{yu2025dyn} + MV-SAM3D~\cite{Baicheng_2025} & 5.81 & 1min \\
        \textcolor{black}{POEM~\cite{yang2025multi} + MV-SAM3D~\cite{Baicheng_2025}} & \textcolor{black}{5.30} & \textcolor{black}{1min} \\
        \midrule
        HOST (Ours) & 3.07 & 2s \\
        Full Pipeline (Ours) & \textbf{2.35} & 0.5min \\
        \bottomrule
    \end{tabular}
    \vspace{-5mm}
\end{table}

We first evaluate feedforward approaches by comparing our HOST results against state-of-the-art pose estimators under a sparse 6-view setup. 
We compare against HORT~\cite{chen2025hort}, a composite baseline comprising DynHaMR~\cite{yu2025dyn} for hands and MV-SAM3D~\cite{Baicheng_2025} for objects, \textcolor{black}{and the multi-view hand reconstruction framework POEM~\cite{yang2025multi} combined with MV-SAM3D}. 
To establish an optimally competitive multi-view baseline for these methods, we provide them with significant informational advantages. 
For the inherently monocular hand trackers (HORT and DynHaMR), we elevate their predictions into metric space using ground-truth camera parameters, aggregating their root-relative articulations via a robust 3D spatial median across all 6 views followed by a rigid registration step. 
For the object reconstruction baseline, we utilize MV-SAM3D, an open-source multi-view extension. To ensure a highly competitive evaluation, we explicitly provide it with ground-truth camera parameters and depth maps. 
This allows the model to reliably determine cross-view visibility, enabling it to effectively handle occlusions.

Despite arming these baselines with optimal 3D lifting and pseudo-ground-truth depth, our HOST module outperforms them end-to-end (achieving 3.07 mm vs. 5.81 mm for the monocular-lifted baseline, and \textcolor{black}{5.30 mm for the multi-view POEM baseline}). \textcolor{black}{This performance gap stems from the fact that while POEM leverages multi-view cues, it relies on static frame-by-frame fusion. In contrast,} our factorized spatiotemporal attention effectively leverages \textcolor{black}{continuous} 4D context to maintain accurate and stable trajectories, whereas \textcolor{black}{static} multi-view trackers can still struggle with drift \textcolor{black}{and temporal inconsistency during complex interactions}. 
Building upon this robust initialization, our HOPG module successfully refines the dense geometry, further reducing the surface error to 2.35 mm.

\begin{figure}[t]
    \centering
    \includegraphics[width=\textwidth]{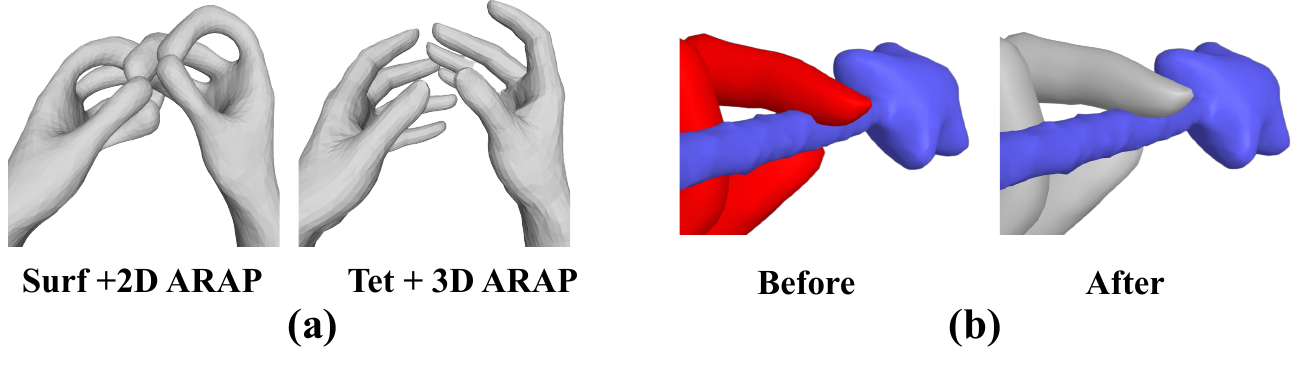}
    \vspace{-9mm}
    \caption{\textbf{Ablation studies.} \textbf{(a) Geometry:} Our tetrahedral formulation (Tet + 3D ARAP) prevents the severe geometric flattening and volume loss seen in surface-only baselines. \textbf{(b) Collision:} Our offline collision optimization successfully resolves non-physical inter-penetrations. }
    \label{fig:ablation_comp}
    \vspace{-7mm}
\end{figure}

\subsection{Ablation Studies}

We investigate the impact of our core design components on our Custom-Captured Dataset.
A comprehensive analysis of our physical constraints and collision optimization is presented in Fig.~\ref{fig:ablation_comp}.

\noindent\textbf{Physics-Aware Geometry Constraint.} 
We evaluate the necessity of our tetrahedral representation (Tet + 3D ARAP) compared to a standard surface-based approach (Surf + 2D ARAP). 
Since our pipeline operates entirely without ground-truth segmentation masks, optimizing exclusively via photometric rendering loss exposes the critical weakness of 2D ARAP: its inability to constrain internal volume, as illustrated in Fig.~\ref{fig:ablation_comp}(a).
In contrast, computing the ARAP energy over the 3D tetrahedral domain explicitly locks the internal structure. This rigorously preserves anatomical thickness, successfully preventing both texture-driven flattening and background-induced geometric collapse.

\noindent\textbf{Collision Optimization.} 
While photometric losses ensure visual alignment, they cannot constrain invisible regions. As shown in Fig.~\ref{fig:ablation_comp}(b), our offline physics-based collision penalty is essential to resolve non-physical inter-penetrations, yielding plausible contact mechanics without sacrificing rendering quality.

\noindent\textbf{Appearance Decoupling.} As shown in Fig.~\ref{fig:main_gallery}, our physics-based model successfully factorizes the rendering into diffuse and specular components. Explicitly modeling view-dependent highlights prevents lighting artifacts from erroneously baking into the geometry. While standard 3DGS frequently warps surfaces or creates unnatural bumps to fake specular glints, our strict disentanglement eliminates these structural hallucinations. This yields pristine, noise-free normal maps, providing the geometric and photometric purity essential for extracting high-quality, relightable 4D digital assets.

\noindent\textbf{Number of Camera Views.} 
We also evaluate the system's performance under varying numbers of input camera views. In this ablation, the view restriction is strictly applied end-to-end. As shown in Fig.~\ref{fig:hand26m_camera}(b), reconstruction quality scales predictably with the camera count. Notably, even in a sparse 4-view setup, the strong spatiotemporal priors from HOST prevent the optimization from collapsing, maintaining a PSNR of nearly 27.89 dB. This confirms that our framework remains highly robust under severe view scarcity.

\textcolor{black}{\section{Limitations}}
\textcolor{black}{While our framework achieves high-fidelity 4D reconstruction, it operates under certain limitations. First, our system relies on synchronized and calibrated multi-view camera setups, which constrains its immediate deployment to unconstrained daily environments. Second, our pipeline currently assumes rigid objects, which prevents it from handling highly deformable items during interactions. Lastly, although our spatiotemporal prior is highly robust, systematic failure cases can still emerge under challenging scenarios: we observed tracking instabilities on geometrically thin or tiny objects (e.g., coins), and extreme motion blur or prolonged cross-hand occlusions could potentially lead to trajectory drift. Future work will integrate non-rigid deformation fields and diversify the training corpus to transition towards uncalibrated, in-the-wild captures.}

\section{Conclusion}
We present a unified framework that bridges accurate kinematic tracking and photorealistic novel view synthesis for 4D hand-object interactions. By coupling a Hand Object Spatiotemporal Transformer (HOST) with a Physics-aware Gaussian (HOPG) module, our method robustly resolves the severe geometric flattening and volume loss common in unconstrained rendering. Furthermore, explicitly decoupling diffuse and specular reflections yields clean, noise-free normal maps, protecting the underlying 3D assets from baked-in lighting artifacts. \textcolor{black}{By automating the workflow from multi-view video to high-fidelity digital assets, this framework offers an efficient solution for large-scale 4D asset creation.}


%
%
\bibliographystyle{splncs04}
\bibliography{main}

\vfill\eject 
\appendix
\section*{Supplementary Material}
\renewcommand{\thesection}{\Alph{section}} 

This supplementary document provides additional details omitted from the main text due to space constraints. 
It is organized into two primary sections: Section~\ref{sec:ablations} presents extended ablation studies. 
Section~\ref{sec:method_training} provides comprehensive implementation and training details, including the full formulation of our HOST, multi-task loss functions, and hyperparameter schedules for both the HOST and HOPG modules. 
Additionally, a supplementary demo video is provided to showcase our results. 
The video specifically contrasts the robust initialization provided by the HOST model with the final high-fidelity 4D reconstructions.

\section{Additional Ablation Studies}
\label{sec:ablations}

\subsection{HOST: Network Architecture and Training Data}
To isolate the impact of our architectural designs and training data, we conduct an ablation study on the raw network predictions, bypassing downstream refinements such as Pose Graph Optimization (PGO) and ICP. We evaluate performance using direct regression metrics: Hand Vertex Reconstruction Error (MSE between vertices derived from predicted and ground-truth MANO parameters), Object Rotation Error (Geodesic), and Object Translation Error (MSE). This evaluation is performed on a reserved validation set of 200 non-synthetic samples (5 frames $\times$ 6 views), with results summarized in Table~\ref{tab:sthot_ablation}.

\begin{table}[h]
    \centering
    \caption{\textbf{Ablation Study on HOST Architecture and Data.} Performance of core modules and data components using direct regression metrics on raw network outputs. All MSE metrics are scaled by $10^3$ for readability.}
    \label{tab:sthot_ablation}
    \resizebox{\linewidth}{!}{%
    \begin{tabular}{lccc}
        \toprule
        \textbf{Model Configuration} & \textbf{Hand Vertices $\downarrow$} & \textbf{Obj. Rot. $\downarrow$} & \textbf{Obj. Trans. $\downarrow$} \\
        & {\small (MSE $\times 10^3$)} & {\small (Geodesic)} & {\small (MSE $\times 10^3$)} \\
        \midrule
        (a) Naive VGGT Attn. (Frame-Global) & 6.65  & 0.0928 & 55.10 \\
        (b) w/o Camera Embedding            & 7.12  & 0.0933 & 64.30 \\
        (c) w/o Time Token                  & 7.36  & 0.1011 & 74.10 \\
        (d) w/o Dense Geometry Heads        & 6.86  & 0.0939 & 65.20 \\
        \midrule
        (e) w/o Synthetic Data Training     & 7.28  & 0.2849 & 87.80 \\
        (f) w/o HOPG Pseudo-GT              & 8.06  & 0.1135 & 78.50 \\
        \midrule
        \textbf{HOST (Full Model)}          & \textbf{6.64} & \textbf{0.0901} & \textbf{54.50} \\
        \bottomrule
    \end{tabular}%
    }
\end{table}

\noindent\textbf{Architectural Designs.} 
Replacing our factorized spatiotemporal attention with the naive frame-global attention from VGGT (Row a) yields comparable regression accuracy but significantly increases computational complexity and memory footprint, limiting efficient multi-view processing. 
The inclusion of camera embeddings (Row b) and time tokens (Row c) is essential for the network to resolve spatiotemporal cues. Specifically, when time tokens are omitted, the model predicts hand and object states by averaging visual image tokens, which leads to a decrease in both localization accuracy and temporal consistency.
Furthermore, although removing the dense geometry heads (row d) only results in a slight decrease in the raw regression metrics, these heads are necessary for providing the 3D spatial priors (e.g. depth and masks) needed for the Pose Graph Optimization and ICP refinement stages.

\noindent\textbf{Training Data Distribution.} 
Rows (e) and (f) evaluate the impact of our training data. Excluding the synthetic object dataset (Row e) reduces the network's ability to generalize for 6D rotation estimation on unseen objects. 
Additionally, fine-tuning with HOPG-refined pseudo-ground truth (Row f) is necessary to bridge the domain gap.

\begin{figure}[t]
    \centering
    \includegraphics[width=\textwidth]{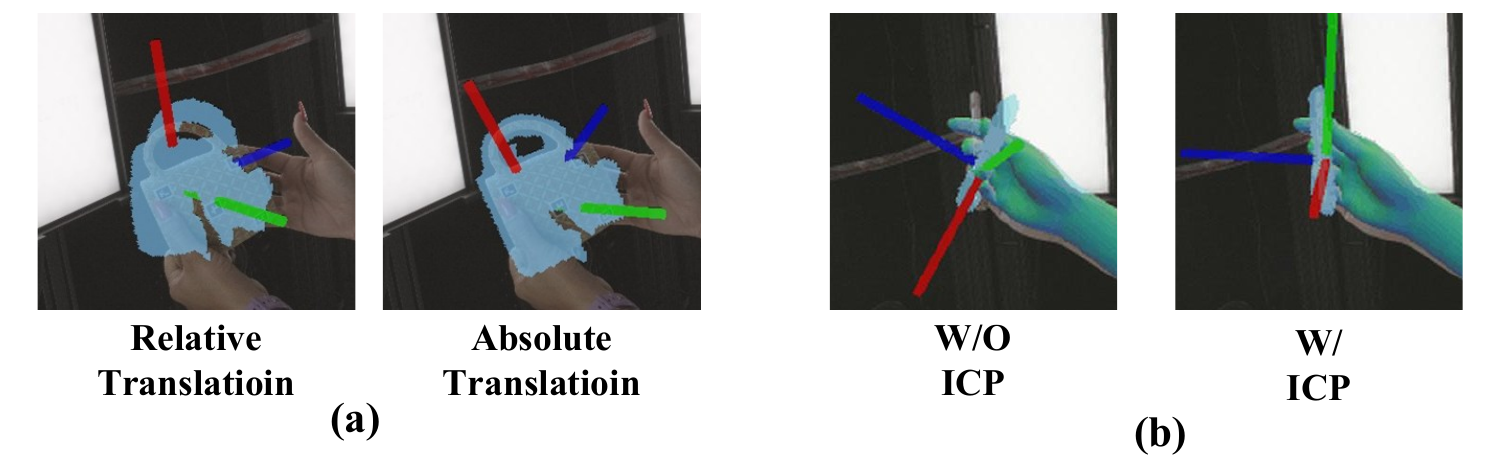}
    \vspace{-9mm}
    \caption{\textbf{Qualitative Ablation on Kinematic Representation and ICP Refinement.} 
    (Left) Compared to relative offsets, our absolute centroid regression provides a global anchor that eliminates cumulative translational drift. 
    (Right) PGO aggregates short-window predictions but is susceptible to rotational drift over long sequences. Our ICP stage uses predicted geometric priors as absolute anchors to correct these offsets and ensure long-term pose consistency.}
    \label{fig:ablation_drift_icp}
\end{figure}

\subsection{HOST: Translation Representation}
As discussed in the main paper, we use an asymmetric kinematic representation of relative rotation combined with absolute centroid translation. We validate this design by comparing it to a fully relative representation. As illustrated in Fig.~\ref{fig:ablation_drift_icp} (left), relative translation inevitably leads to cumulative error and severe positional drift over long sequences. In contrast, regressing the absolute geometric center provides a global anchor for the trajectory. This ensures that translation errors remain bounded and consistent, regardless of sequence duration.

\subsection{HOST: ICP Correction}
Although our PGO module successfully combines the predicted relative rotations of the short-window transformer into a continuous trajectory, the resulting global poses remain susceptible to cumulative rotational drift over long sequences. 
To address this issue, we introduce an ICP-based refinement stage that uses the dense geometry priors (depth and masks) predicted by our dense heads. Aligning the object point cloud to these per-frame spatial anchors provides an absolute geometric reference through ICP correction, which rectifies the accumulated drift. As illustrated in Fig.~\ref{fig:ablation_drift_icp} (right), this step ensures long-term pose consistency.

\section{Method and Implementation Details}
\label{sec:method_training}

In this section, we provide comprehensive architectural details, loss formulations, and training hyperparameters for both the Hand Object Spatiotemporal Transformer (HOST) and the Hand Object Physics-aware Gaussian (HOPG) refinement module.

\subsection{Hand Object Spatiotemporal Transformer (HOST)}

\begin{figure*}[htbp]
\centering
\resizebox{\linewidth}{!}{%
\begin{tikzpicture}[
    font=\sffamily\scriptsize,
    >=Stealth,
    node distance=0.5cm and 0.8cm,
    input/.style={rectangle, draw=black!80, thick, fill=gray!10, text width=1.8cm, text centered, rounded corners, minimum height=0.8cm},
    embed/.style={rectangle, draw=black!80, thick, fill=blue!5, text width=2.4cm, text centered, rounded corners, minimum height=0.8cm},
    attn/.style={rectangle, draw=black!80, thick, fill=orange!10, text width=2.5cm, text centered, rounded corners, minimum height=0.5cm},
    head/.style={rectangle, draw=black!80, thick, fill=green!10, text width=2.4cm, text centered, rounded corners, minimum height=0.8cm},
    output/.style={rectangle, draw=black!80, thick, fill=red!10, text width=1.8cm, text centered, rounded corners, minimum height=0.6cm},
    op/.style={circle, draw=black!80, thick, fill=yellow!20, inner sep=2pt, font=\bfseries\large},
    tensor/.style={text=black, font=\ttfamily\bfseries\fontsize{6}{7}\selectfont, fill=white, inner sep=1pt},
    dashbox/.style={draw=black!50, dashed, thick, inner sep=8pt, rounded corners, fill=blue!2},
    tokenbox/.style={rectangle, draw=blue!60, thick, fill=blue!10, text width=2.2cm, text centered, rounded corners, minimum height=0.8cm}
]

\node (img) [input] {Input Images\\$(B, S, 3, H, W)$};
\node (cam) [input, above=0.3cm of img] {Camera Params};
\node (fids) [input, below=0.3cm of img] {Relative\\Frame IDs};

\node (dinov2) [embed, right=of img] {Frozen DINOv2\\Patch Embed};
\node (cam_embed) [embed, right=of cam] {Camera Encoders\\(MapAnything Style)};
\node (time_embed) [embed, right=of fids] {Time Embedding};

\draw [->, thick] (img) -- (dinov2);
\draw [->, thick] (cam) -- (cam_embed);
\draw [->, thick] (fids) -- (time_embed);

\node (concat) [op, right=0.6cm of dinov2] {+};
\node (tokens) [embed, right=0.6cm of concat, text width=2cm] {Tokens Prep\\(+ Register)};

\draw [->, thick] (dinov2) -- (concat);
\draw [->, thick] (cam_embed) -| (concat);
\draw [->, thick] (concat) -- (tokens);
\draw [->, thick] (time_embed.east) -| (tokens.south); 

\node (spatial) [attn, right=2.2cm of tokens, yshift=0.45cm] {\textbf{Spatial Attn}};
\node (frame2) [attn, right=2.2cm of tokens, yshift=-0.45cm] {\textbf{Frame Attn}};
\node (frame1) [attn, above=0.5cm of spatial] {\textbf{Frame Attn}};
\node (temporal) [attn, below=0.5cm of frame2] {\textbf{Temporal Attn}};

\begin{scope}[on background layer]
    \node (attn_box) [dashbox, fit=(frame1) (spatial) (frame2) (temporal)] {};
    \node [anchor=south, font=\bfseries] at (attn_box.north) {$12\times$ Spatiotemporal Blocks};
\end{scope}

\draw [->, thick] (tokens) -- (attn_box.west) node[midway, above, tensor] {$[B*S, P, C]$};

\draw [->, thick] (frame1) -- (spatial) node[midway, right, tensor] {$[B*F, V*P, C]$};
\draw [->, thick] (spatial) -- (frame2) node[midway, right, tensor] {$[B*S, P, C]$};
\draw [->, thick] (frame2) -- (temporal) node[midway, right, tensor] {$[B*V, F*P, C]$};

\node (split) [coordinate, right=1.2cm of attn_box.east] {};
\draw [->, thick] (attn_box.east) -- (split) node[midway, above, tensor] {$[B, S, P, C]$};

\node (dpt) [head, above right=0.8cm and 0.8cm of split, text width=2.8cm] {Multi-scale DPTs\\(Layers 2, 5, 8, 11)};
\draw [->, thick] (split) |- (dpt);

\node (out_point) [output, right=1cm of dpt] {Point};
\node (out_depth) [output, above=0.2cm of out_point] {Depth};
\node (out_seg) [output, below=0.2cm of out_point] {Seg Mask};

\draw [->, thick] (dpt.east) -- ++(0.4,0) |- (out_depth.west);
\draw [->, thick] (dpt.east) -- (out_point.west);
\draw [->, thick] (dpt.east) -- ++(0.4,0) |- (out_seg.west);

\node (obj_head) [head, below right=0.3cm and 0.8cm of split] {Object Pose Head};
\node (hand_head) [head, below=0.4cm of obj_head] {Hand Pose Head};

\draw [->, thick] (split) |- (obj_head);
\draw [->, thick] (split) |- (hand_head);

\node (out_obj) [output, right=1cm of obj_head] {Object State};
\node (out_hand) [output, right=1cm of hand_head] {Hand State};

\draw [->, thick] (obj_head) -- (out_obj);
\draw [->, thick] (hand_head) -- (out_hand);

\node (detail_bg) [draw=black!40, dashed, thick, fill=green!5, rounded corners, minimum width=15.2cm, minimum height=2.5cm, below=0.8cm of temporal, xshift=0.8cm] {};
\node [anchor=north west, font=\bfseries] at (detail_bg.north west) {Zoom-in: Hand Pose Head Details};

\node (detail_in) [tokenbox, right=0.3cm of detail_bg.west] {\fontsize{5}{6}\selectfont Updated Time Tokens\\$[B, F, V, C]$};

\node (detail_attn) [attn, right=0.7cm of detail_in, text width=2.6cm] {View-Adaptive Attn\\(Pool over $V$)};
\node (detail_trunk) [attn, fill=blue!10, right=0.8cm of detail_attn, text width=2.6cm] {Temporal Trunk\\(Self-Attn over $F$)};
\node (detail_mlp) [head, right=0.8cm of detail_trunk, text width=1.2cm] {MLPs};

\node (detail_out_global) [output, right=0.8cm of detail_mlp] {Global Pose};
\node (detail_out_shape) [output, above=0.2cm of detail_out_global] {Shape};
\node (detail_out_local) [output, below=0.2cm of detail_out_global] {Joint Pose};

\draw [->, thick] (detail_in) -- (detail_attn);
\draw [->, thick] (detail_attn) -- (detail_trunk) node[midway, above, tensor, fill=green!5, inner sep=1pt] {$[B, F, C]$};
\draw [->, thick] (detail_trunk) -- (detail_mlp) node[midway, above, tensor, fill=green!5, inner sep=1pt] {$[B, F, C]$};
\draw [->, thick] (detail_mlp.east) -- ++(0.4,0) |- (detail_out_shape.west);
\draw [->, thick] (detail_mlp.east) -- (detail_out_global.west);
\draw [->, thick] (detail_mlp.east) -- ++(0.4,0) |- (detail_out_local.west);

\draw [->, dashed, thick, draw=black!50] (hand_head.south) -- (detail_bg.north-|hand_head.south);

\end{tikzpicture}
}
\vspace{2mm}
\caption{\textbf{Detailed Architecture of HOST.} The network extracts features using a frozen DINOv2 backbone integrated with camera and time embeddings. Given an input sequence of $F$ frames from $V$ views ($S = F \times V$ images total), 12 spatiotemporal blocks process the tokens by alternating between spatial (cross-view) and temporal (cross-frame) attention. These shared feature representations are then fed into task-specific heads: a dense geometry decoder for depth, point, and mask prediction, and parametric heads for hand and object pose regression. In the pose heads (bottom), updated time tokens from all $V$ views are adaptively pooled into canonical per-frame features, which are then processed by a temporal trunk to ensure motion continuity prior to kinematic parameter regression.}
\label{fig:sthot_arch}
\end{figure*}
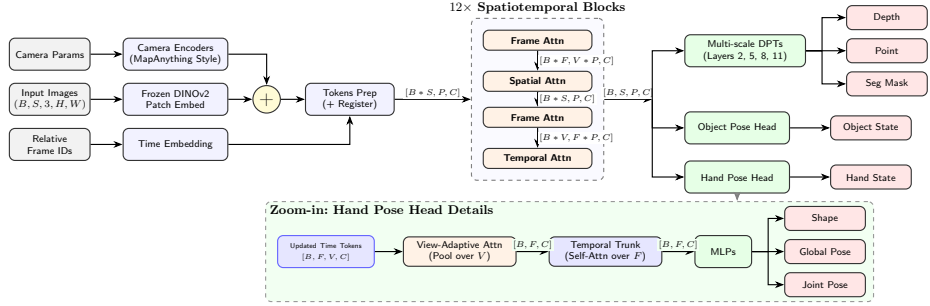

\noindent\textbf{Patch and Camera Embeddings.} We process $308 \times 308$ images using a frozen DINOv2~\cite{oquab2023dinov2} (ViT-L/14) patch embedding layer. To inject geometric and temporal awareness, we follow MapAnything~\cite{keetha2025mapanything} to augment each patch token with ray-based camera embeddings, which encode camera geometry via 4-layer MLPs. 
For temporal information, we map relative frame IDs through a learnable time embedding layer to obtain frame-specific time tokens. These time tokens are concatenated with the image tokens and 4 register tokens~\cite{wang2025vggt} before being fed into the spatiotemporal aggregator.

\noindent\textbf{Spatiotemporal Attention.} The core aggregator consists of $L=12$ Transformer blocks. 
As detailed in the main paper, we decompose the full global attention into a sequential factorized scheme (Frame $\rightarrow$ Spatial $\rightarrow$ Frame $\rightarrow$ Temporal) to maintain computational tractability. 
In our implementation, each sub-attention layer retains the same hidden dimension $C=1024$ as the VGGT backbone~\cite{wang2025vggt}. 
To effectively leverage pre-trained geometric priors, we initialize the aggregator by mapping weights from VGGT: the frame attention weights are directly transferred to our frame attention layers, while the weights from VGGT's global attention layers are sequentially mapped to initialize our spatial and temporal attention blocks.

\noindent\textbf{Task-Specific Heads.}  
For dense predictions including depth, masks, and point maps, we employ the DPT-based decoders~\cite{ranftl2021vision} following VGGT~\cite{wang2025vggt}, which leverage multi-scale features from intermediate layers $\{2, 5, 8, 11\}$. 

For kinematic states, we first extract the updated time tokens $\mathbf{\hat{T}}^{f,v}_{\text{time}}$ from all views. 
To resolve occlusions and compute robust frame-level features $\mathbf{z}_f$, we introduce a view-adaptive attention mechanism applied separately to each entity. 
For the hand or object, a dedicated MLP maps the corresponding time tokens to softmax-normalized visibility weights $\alpha_{f,v}$ across all $V$ views:
\begin{equation}
    \alpha_{f,v} = \frac{\exp(\text{MLP}_{\text{attn}}(\mathbf{\hat{T}}^{f,v}_{\text{time}}))}{\sum_{k=1}^{V} \exp(\text{MLP}_{\text{attn}}(\mathbf{\hat{T}}^{f,k}_{\text{time}}))}
\end{equation}
The resulting aggregated feature $\mathbf{z}_f = \sum_{v=1}^{V} \alpha_{f,v} \mathbf{\hat{T}}^{f,v}_{\text{time}}$ is then processed by a separate temporal trunk—consisting of self-attention blocks over the temporal dimension $F$—to ensure motion smoothness and continuity for each entity. 
Finally, dedicated MLP decoders regress these temporally smoothed features into kinematic parameters. Specifically, the object head predicts absolute 3D translations and relative 6D rotations, while the hand head regresses 10D shape parameters, 9D global poses, and 63D local joint Euler angles~\cite{MANO:SIGGRAPHASIA:2017}.

\subsubsection{Loss Formulation}
The HOST is trained end-to-end using a multi-task loss:
\begin{equation}
    \mathcal{L}_{total} = \lambda_{depth}\mathcal{L}_{depth} + \lambda_{point}\mathcal{L}_{point} + \lambda_{mask}\mathcal{L}_{mask} + \lambda_{obj}\mathcal{L}_{obj} + \lambda_{hand}\mathcal{L}_{hand}
\end{equation}
During the joint training stage, we set $\lambda_{depth}=0.5$, $\lambda_{point}=0.5$, $\lambda_{mask}=1.0$, $\lambda_{obj}=1.0$, and $\lambda_{hand}=1.0$. 

\noindent\textbf{Dense Losses.} We supervise segmentation masks with a standard binary cross-entropy loss. For depth and 3D point maps, we adopt the same loss formulation as VGGT~\cite{wang2025vggt}.

\noindent\textbf{Kinematic Losses.} We employ the geodesic loss for all 3D rotations:
\begin{equation}
    \mathcal{L}_{geo}(\mathbf{R}_{pred}, \mathbf{R}_{gt}) = \arccos\left(\frac{\text{tr}(\mathbf{R}_{pred}\mathbf{R}_{gt}^T) - 1}{2}\right)
\end{equation}
The object pose is supervised through global translation and rotation errors:
\begin{equation}
    \mathcal{L}_{obj} = \lambda_{trans}\|\mathbf{t}_{pred} - \mathbf{t}_{gt}\|_2 + \lambda_{rot}\mathcal{L}_{geo}(\mathbf{R}_{pred}, \mathbf{R}_{gt})
\end{equation}
To accurately recover the articulated hand, we jointly supervise the global pose, local joint rotations, shape parameters, and the resulting mesh vertices.
\begin{align}
\begin{split}
    \mathcal{L}_{hand} &= \lambda_{trans}\|\mathbf{t}_{pred} - \mathbf{t}_{gt}\|_2 + \lambda_{rot}\mathcal{L}_{geo}(\mathbf{R}_{pred}, \mathbf{R}_{gt}) \\
    &\quad + \lambda_{joint}\mathcal{L}_{geo}(\mathbf{R}^{joint}_{pred}, \mathbf{R}^{joint}_{gt}) + \lambda_{shape} \| \mathbf{\alpha}_{pred} - \mathbf{\alpha}_{gt} \|_2 \\
    &\quad + \lambda_{vert} \|\mathbf{V}_{pred} - \mathbf{V}_{gt}\|_2
\end{split}
\end{align}
For object and hand loss, we set the loss weights to $\lambda_{trans} = 5.0$ and $\lambda_{vert} = 5.0$, while all other coefficients are set to $1.0$.

\subsubsection{Training Strategy and Hyperparameters}
HOST is implemented in JAX/Flax and optimized using the AdamW optimizer in bfloat16 precision. 
Given that the backbone is initialized with robust pre-trained weights, we apply a learning rate multiplier of $5.0$ to the newly added mask, hand, and object heads to harmonize their optimization with the deep backbone.
The training pipeline is structured into two stages:

\noindent\textbf{Stage 1: Large-scale Pre-training.} We initialize our network using pre-trained weights from VGGT~\cite{wang2025vggt} to leverage robust geometric priors. 
Specifically, we load weights for the DINO backbone, attention aggregator, depth head, and point head directly. 
The model is trained on 64 TPU v5 accelerators for 200,000 steps with a total batch size of 64, using the large-scale VEPHand dataset~\cite{anonymous_handbooth} introduced in the main paper. 
The base learning rate is $8.0 \times 10^{-5}$ with a 5,000-step linear warmup, followed by a piecewise constant decay by factors of $\{0.5, 0.2, 0.1\}$ at steps $\{60\text{k}, 80\text{k}, 120\text{k}\}$.

\noindent\textbf{Stage 2: Fine-tuning.} 
The network is fine-tuned for 200,000 steps on 8 NVIDIA H100 GPUs with a total batch size of 16, using a data mixture of the VEPHand dataset, synthetic object dataset~\cite{zhang2025texverse} and pseudo-ground-truth sequences refined by HOPG. 
The base learning rate is reduced to $1.0 \times 10^{-5}$ with a 5,000-step linear warmup to adapt to the target data distribution.

\noindent\textbf{Geometry Warmup.} At the beginning of both training stages, we enforce a geometry warmup for the initial 2,000 steps by setting $\lambda_{obj} = \lambda_{hand} = 0.0$. 
This allows the dense geometric features to stabilize before regressing kinematic poses.

\subsection{Hand Object Physics-aware Gaussian (HOPG) Details}

\subsubsection{Mesh-Aligned 2D Gaussian Parameterization}
To reconstruct high-fidelity surfaces while preserving kinematic consistency, we anchor 2D Gaussians directly to the hand template meshes. Following the barycentric parameterization in PICA~\cite{peng2025pica}, we extend the formulation by employing a hybrid sampling strategy that anchors Gaussians to both triangular faces and mesh vertices. 
For each vertex-anchored Gaussian, the normal vector $\mathbf{v}_0$ is derived from the vertex normal, while the tangent vectors $\mathbf{v}_1$ and $\mathbf{v}_2$ are determined via Gram-Schmidt orthogonalization against a reference axis. 
This hybrid parameterization ensures that the 2D Gaussians remain strictly coupled with the deforming mesh, preventing spatial scattering and ensuring surface integrity during articulated motion.

\subsubsection{Pose-Conditioned Volumetric Deformation}
As described in the main paper, we predict the non-rigid displacements for the tetrahedral vertices $\boldsymbol{\rho}_c$ using a lightweight MLP consisting of 5 layers with a hidden dimension of 256. This network takes both the canonical coordinates $\boldsymbol{\rho}_c$ and the 63-dimensional axis-angle pose $\beta$ as input to generate pose-dependent volumetric offsets. To ensure numerical stability, the final linear layer is initialized from a uniform distribution bounded by $[-10^{-5}, 10^{-5}]$. This initialization yields near-zero initial offsets, allowing the as-rigid-as-possible (ARAP) energy and Linear Blend Skinning (LBS) to provide a stable geometric starting point before the model learns fine-grained non-rigid deformations.

\subsubsection{Normal-Depth Consistency}
To enhance surface smoothness and geometric accuracy, we employ the normal-depth consistency loss as proposed in 2DGS~\cite{huang20242d}. Specifically, we enforce an L1 alignment between the rendered normals and the pseudo-ground-truth normals derived from the spatial gradients of the depth map. Rather than using standard median depth, we utilize the depth $Z_{pgsr}$ generated via Plane-based Gaussian Splatting (PGSR)~\cite{chen2024pgsr}, which provides more robust geometric constraints:
\begin{equation}
    Z_{pgsr} = \frac{\sum \alpha_i d_i}{\sum \alpha_i |\mathbf{n}_i \cdot \mathbf{r}| + \epsilon}
\end{equation}
As in the original 2DGS formulation, we apply a soft-blending mask to ignore pixels with grazing viewing angles, ensuring stable gradient propagation for the consistency loss.

\subsubsection{ARAP and Offline Collision Optimization}
To prevent internal volume collapse during non-rigid deformation, we enforce an As-Rigid-As-Possible (ARAP) energy on the tetrahedral mesh. The energy is computed between the canonical tetrahedral vertices $\boldsymbol{\rho}_c$ and the deformed vertices $\boldsymbol{\rho}'_c$:
\begin{equation}
    \mathcal{L}_{arap} = \sum_{i} \sum_{j \in \mathcal{N}(i)} w_{ij} \left\| (\boldsymbol{\rho}'_{c,i} - \boldsymbol{\rho}'_{c,j}) - \mathbf{R}_i (\boldsymbol{\rho}_{c,i} - \boldsymbol{\rho}_{c,j}) \right\|_2^2
\end{equation}
where $\mathbf{R}_i$ is the optimal rotation matrix estimated via Singular Value Decomposition (SVD) for the local neighborhood of the $i$-th tetrahedral cell. 

\noindent\textbf{Offline Collision Resolution.} Since photometric supervision alone may not fully resolve physical penetrations, we incorporate a physics-aware refinement stage to enforce contact constraints. We categorize collisions into \textit{self-collision} (intra-hand) and \textit{inter-collision} (hand-object). Specifically, for a set of query vertices $\{\mathbf{p}_q\}$, we first utilize a high-performance CUDA kernel to perform a point-in-mesh containment check, generating a binary mask $\mathcal{M}$ where $\mathcal{M}_q=1$ for vertices situated inside the reference geometry. For each identified penetration, we compute its nearest reference surface point $\mathbf{p}_{ref}$ and minimize the distance to the surface projected along the query vertex's normal $\mathbf{n}_q$:
\begin{equation}
    \mathcal{L}_{coll} = \sum_{q} \mathcal{M}_q \cdot \left( (\mathbf{p}_q - \mathbf{p}_{ref}) \cdot \mathbf{n}_q \right)^2
\end{equation}
Every 15,000 global training steps, we trigger a dedicated offline optimization phase. During this phase, we perform 1,000 iterations using the Adam optimizer to update the hand pose parameters $\Delta \beta$ exclusively, thereby effectively resolving any accumulated intersections.

\subsubsection{Loss Functions and Optimization Details}
To explicitly decouple environmental lighting parameters from the intrinsic appearance of Gaussian primitives, the HOPG module is optimized using a batch size of $N_b=5$. This joint optimization across multiple frames allows the model to differentiate between view-dependent lighting effects and the consistent underlying geometry and texture.
For a typical 21-view, 600-frame sequence, the optimization runs for 30,000 iterations, taking approximately 5 hours on a single NVIDIA A100 GPU. The overall objective function is formulated as:
\begin{align}
\begin{split}
    \mathcal{L}_{HOPG} &= \lambda_{photo}\mathcal{L}_{L1} + \lambda_{dssim}\mathcal{L}_{D-SSIM} + \lambda_{lpips}\mathcal{L}_{LPIPS} \\
    &\quad + \lambda_{arap}\mathcal{L}_{arap} + \lambda_{lap}\mathcal{L}_{lap} + \lambda_{norm}\mathcal{L}_{norm} + \lambda_{smooth}\mathcal{L}_{smooth}
\end{split}
\end{align}

The primary photometric loss weights are set to $\lambda_{photo}=1.0$, $\lambda_{dssim}=0.2$, and $\lambda_{lpips}=2.0$. For geometric regularization, we employ a Tet-ARAP weight of $\lambda_{arap}=5000.0$ and a Laplacian smoothness weight of $\lambda_{lap}=20000.0$ to preserve mesh structural integrity, alongside a PGSR normal consistency weight $\lambda_{norm}=1.0$. Temporal smoothness penalties $\mathcal{L}_{smooth}$ are applied to both hand poses $\theta$ and object trajectories with $\lambda_{smooth}=10.0$ to ensure kinematic continuity.

\section{\textcolor{black}{Computational Efficiency and Scalability}}
\label{sec:efficiency}

\textcolor{black}{As our system targets high-fidelity offline 4D asset reconstruction, we prioritize robustness and accuracy. Below, we detail the computational requirements of our two main stages.}

\begin{table}[h]
\centering
\caption{\textbf{HOST Inference Runtime and Peak Memory} relative to the number of views ($V$) on a single H100 GPU.}
\label{tab:memory}
\vspace{2mm} 

\resizebox{0.6\linewidth}{!}{%
\begin{tabular}{lcccc}
\toprule
Number of Views ($V$) & 4 & 6 & 10 & 19 \\
\midrule
HOST Infer Time (s) & 0.34 & 0.82 & 1.24 & 4.38 \\
Peak VRAM (GB) & $\sim$7.4 & $\sim$11.5 & $\sim$23.8 & $\sim$75.6 \\
\bottomrule
\end{tabular}%
}
\end{table}

\vspace{2mm}
\noindent \textbf{HOST Inference:} With a temporal window $F=5$, HOST's cost comprises model inference (which scales roughly quadratically with the number of views $V$ due to the attention mechanism) and view-agnostic post-processing (PGO+ICP, around 1 frame/s). Table~\ref{tab:memory} details the runtime and peak memory measurements on a single NVIDIA H100 GPU.

\vspace{2mm}
\noindent \textcolor{black}{\textbf{HOPG Optimization:}} \textcolor{black}{For the HOPG refinement stage, the optimization takes approximately 5 hours per sequence on an NVIDIA A100 GPU. This computational cost is standard for high-fidelity, physics-aware offline asset reconstruction. Furthermore, to scale to large datasets, this sequence-level optimization is inherently decoupled and can be trivially distributed across compute clusters in parallel.}

\end{document}